\documentclass[10pt,twocolumn,letterpaper]{article}

\usepackage{iccv}
\usepackage{times}
\usepackage{epsfig}
\usepackage{graphicx}
\usepackage{amsmath}
\usepackage{amssymb}
\usepackage{caption}
\usepackage{subcaption}
\usepackage{bm}
\usepackage{enumitem}

\usepackage[pagebackref=true,breaklinks=true,letterpaper=true,colorlinks,bookmarks=false]{hyperref}

\iccvfinalcopy % *** Uncomment this line for the final submission

\def\iccvPaperID{7975} % *** Enter the ICCV Paper ID here

\ificcvfinal\fi
\begin{document}

%%%%%%%%% TITLE
\title{DeePSD: Automatic Deep Skinning And Pose Space Deformation For 3D Garment Animation}

\author{Hugo Bertiche$^{1,2}$, Meysam Madadi$^{1,2}$, Emilio Tylson$^{1}$ and Sergio Escalera$^{1,2}$\\
$^1$University of Barcelona and $^2$Computer Vision Center, Spain\\
{\tt\small hugo\_bertiche@hotmail.com}
}
%\twocolumn[{%
%\renewcommand\twocolumn[1][]{#1}%
\maketitle
%\begin{center}
%\vspace{-0.8cm}
%    \includegraphics[width=0.48\textwidth]{img/title4.png}
%    \vspace{-0.3cm}
%    \captionof{figure}{We present an approach to the garment animation problem allowing automatic animation of any unseen outfit of arbitrary complexity and topology, including ornaments, through deep learning. This is the first work able to obtain this level of generalization. This heavily increases model applicability in the videogame industry, 3D modelling and virtual try-ons.}
%    \label{fig:title}
%\end{center}%
%}]

%\thispagestyle{empty}

%%%%%%%%% ABSTRACT
\begin{abstract}
We present a novel solution to the garment animation problem through deep learning. Our contribution allows animating any template outfit with arbitrary topology and geometric complexity. Recent works develop models for garment edition, resizing and animation at the same time by leveraging the support body model (encoding garments as body homotopies). This leads to complex engineering solutions that suffer from scalability, applicability and compatibility. By limiting our scope to garment animation only, we are able to propose a simple model that can animate any outfit, independently of its topology, vertex order or connectivity. Our proposed architecture maps outfits to animated 3D models into the standard format for 3D animation (blend weights and blend shapes matrices), automatically providing of compatibility with any graphics engine. We also propose a methodology to complement supervised learning with an unsupervised physically based learning that implicitly solves collisions and enhances cloth quality.\vspace{-1cm}
\end{abstract}

\section{Introduction}
\label{sec:intro}

Virtual dressed human animation has been a topic of interest for decades due to its numerous applications in entertainment and videogame industries, and recently, in virtual and augmented reality. Depending on the application we find two main classical computer graphics approaches. On the one hand, Physically Based Simulation (PBS) \cite{baraff1998large, liu2017quasi, provot1997collision, provot1995deformation, tang2013gpu, vassilev2001fast, zeller2005cloth} approaches are able to obtain highly realistic cloth dynamics at the expense of a huge computational cost. %Also, PBS needs accurate parameter fine-tuning to obtain the desired results, making it a time-consuming task requiring expertise on the field. 
On the other hand, Linear Blend Skinning (LBS) \cite{kavan2008geometric, kavan2005spherical, le2012smooth, Magnenat-thalmann88joint-dependentlocal, wang2007real, wang2002multi} and Pose Space Deformation (PSD) \cite{allen2002articulated, anguelov2005scape, lewis2000pose, loper2015smpl} models are suitable for environments with limited computational resources or real-time performance demand. To do so, realism is highly compromised. Then, computer graphics approaches present a trade-off between realism and computational performance.

Deep learning has already proven successful in complex 3D tasks \cite{arsalan2017synthesizing, han2017deepsketch2face, madadi2020smplr, omran2018neural, qi2017pointnet, richardson20163d, socher2012convolutional}. Due to the interest in the topic and the recently available 3D datasets on garments, we see the scientific community pushing this research line \cite{alldieck2018video, alldieck2019tex2shape, bertiche2020cloth3d, bhatnagar2019multi, guan2012drape, lahner2018deepwrinkles, patel2020tailornet, santesteban2019learning}. Most proposals are built as non-linear PSD models learnt through deep learning. These methods yield models describing one or few garment types and, therefore, they lack on generalization capabilities. To overcome this, recent works propose encoding garment types as a subset of body vertices \cite{bertiche2020cloth3d, patel2020tailornet}. This allows generalizing to more garments, yet bounds its representation capacity to body homotopies only. Thus, these approaches need to model each garment individually and cannot handle details such as pockets nor multiple layers of cloth, heavily hurting their scalability and applicability in real life scenarios.

\begin{figure}[!t]
    \centering
    \includegraphics[width=\columnwidth]{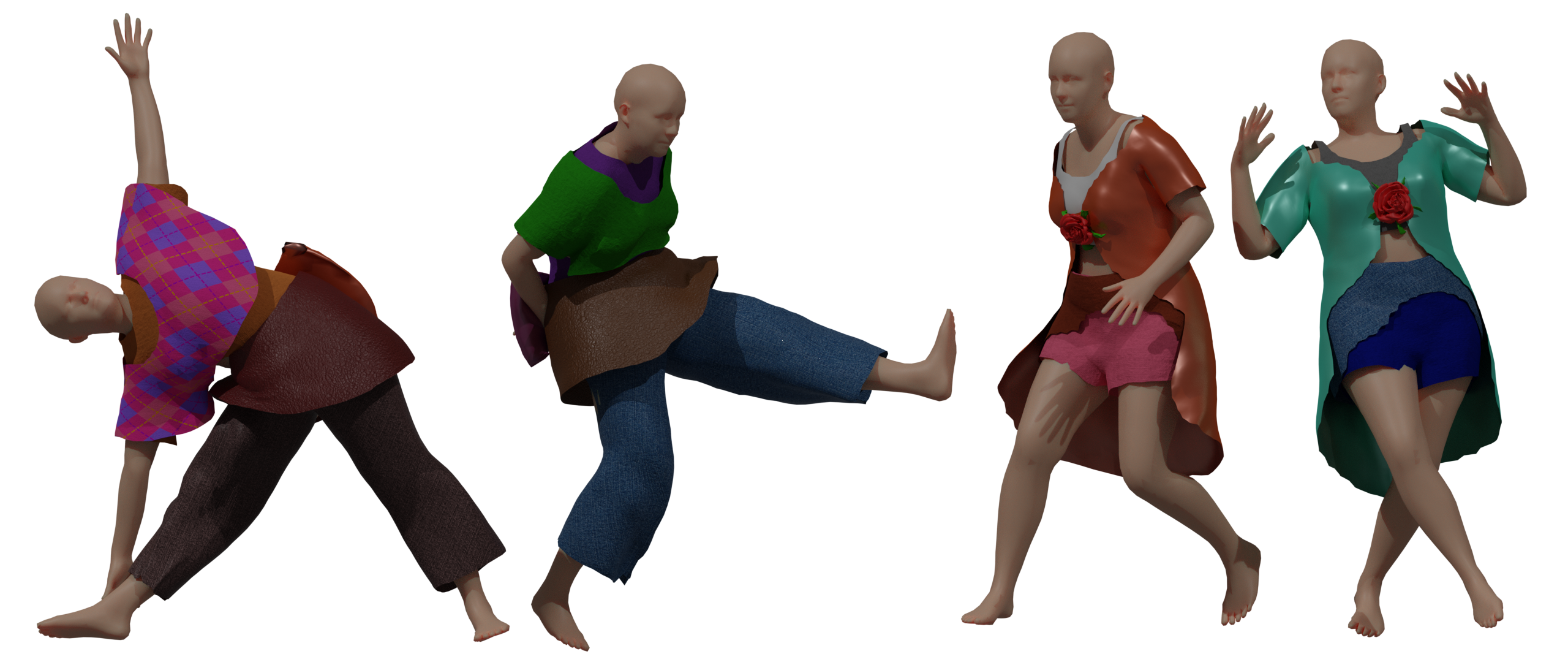}
    \vspace{-.7cm}
    \caption{We present a novel approach for outfit animation. Our methodology allows generalization to unseen outfits. It can handle multiple layers of cloth, arbitrary topology and complex geometric details without retraining. \vspace{-.5cm}}
    \label{fig:title}
\end{figure}

We propose learning a mapping from the space of template outfits to the space of animated 3D models. We will show how this allows generalization to completely unseen garments with arbitrary topology and vertex connectivity. We can achieve this by identifying edition/resizing and animation as separate tasks, and focusing on the latter. % (similarly, the work of \cite{vidaurre2020fully} allows handling garments independently of their vertex order and connectivity by focusing on edition/resizing only). 
Our method works with whole outfits (instead of single garments), multiple layers of cloth and resolutions, while also allowing complex geometric details (see Fig. \ref{fig:title} for some examples). Furthermore, we achieve this with a simple and small-sized neural network. The list of our contributions is as follows:
\begin{itemize}[noitemsep,topsep=0pt,parsep=0pt,partopsep=0pt]
    \item \textbf{Outfit Generalization.} To the best of our knowledge, our proposal is the only work able to animate completely unseen outfits without additional training. This greatly increases applicability in scenarios with ever-growing number of outfits, such as virtual try-ons and videogames, where customization is key.
    %\item \textbf{Automatic Skinning.} 3D models require an skinning w.r.t. an skeleton to be animated. To define valid PSD models, it is required to obtain a skinning consistent w.r.t. the deformations. We are the first to propose an automatic unsupervised deep learning based approach for garment skinning within animation domain. \textcolor{red}{[I'M THINKING ON REMOVING THIS CONTRIBUTION, WE HAVE MANY AND STRONGER ONES.]}
    \item \textbf{Compatibility.} Our methodology does not predict garment vertex locations, but blend weights and blend shapes matrices. This is the standard on 3D animation, and it is therefore compatible with all graphics engines. Also, it benefits from the exhaustive optimization on animation pipelines. Pose Space Deformations are a specific case of blend shapes that are combined consistently with object pose.
    \item \textbf{Physical Consistency.} Related works require a final post-processing step for collision solving. Alternatively, works that train with a collision-solving loss need to find a compromise between physical constraints and vertex error. Thus, predictions still show collisions. We propose to train an independent model branch such that physical consistency losses and supervised losses do not hinder each other. This yields quasi-collision free and cloth-consistent predictions while leveraging the data as much as possible.
    \item \textbf{Explainability.} Mapping outfits to animated 3D models yields a more intuitive work pipeline for CGI artists. Recent works try to address garment resizing/edition along animation by encoding styles into parametric representations \cite{bertiche2020cloth3d,patel2020tailornet}. Thus, expert knowledge is required to obtain the desired results by tuning style parameters.
\end{itemize}

\section{State of the art}
\label{sec:sota}

%The garment animation domain has been widely explored by the computer graphics community.% due to its numerous applications in entertainment industry.
%With the recent developments of deep learning in the 3D domain and the increasingly amount of available data, neural networks begin to show promising results.

\textbf{Computer Graphics.} Obtaining realistic cloth behaviour is possible through \textbf{PBS} (Physically Based Simulation), commonly through the well known \textit{mass-spring} model. Literature on the topic is extensive, focused on improving the efficiency and stability of the simulation by simplifying and/or specializing on specific setups \cite{baraff1998large, provot1997collision, provot1995deformation, vassilev2001fast}, or proposing new energy-based algorithms to enhance robustness, realism and generalization to other soft bodies \cite{liu2017quasi}. Other works propose leveraging the parallel computational capabilities of modern GPUs \cite{tang2013gpu, zeller2005cloth}. These approaches achieve high realism at the expense of a great computational cost. Thus, PBS is not an appropriate solution when real-time performance is required or computational capacity is limited (e.g. in portable devices). On the other hand, for applications that prioritize performance, \textbf{LBS} (Linear Blend Skinning) is the standard approach on computer graphics for animation of 3D models. Each vertex of the object to animate is attached to a skeleton through a set of blend weights that are used to linearly combine joint transformations. In garment domain, outfits are attached to the skeleton driving body motion. This approach has also been widely studied \cite{kavan2008geometric, kavan2005spherical, le2012smooth, Magnenat-thalmann88joint-dependentlocal, wang2007real, wang2002multi}. While it is possible to achieve real-time performance, cloth dynamics are highly non-linear, which results in a significant loss of realism when applied to garments. %Finally, we can also find \textbf{hybrid} approaches in the industry, where tighter parts of the outfit are rigged to a skeleton and loose parts are simulated in a simplified manner and low vertex count. The garment interacts only with the body, not with the environment. It is common to find this approach in modern video games, as it increases realism without excessively hurting performance.

\textbf{Learning-Based.} Due to the drawbacks found in the classical LBS approach, \textbf{PSD} (Pose Space Deformation) models appeared \cite{lewis2000pose}. To avoid artifacts due to skinning, corrective deformations are applied to the mesh in rest pose. Additionally, PSD handles pose-dependant high frequency details of 3D objects. While hand-crafted PSD is possible, in practice, it is learnt from data. We find applications of this technique for body models \cite{allen2002articulated, anguelov2005scape, loper2015smpl}, where deformation bases are computed through linear decomposition of registered body scans. Similarly, in garment domain, Guan et al. \cite{guan2012drape} apply the same techniques for a few template garments on data obtained through simulation. L\"ahner et al. \cite{lahner2018deepwrinkles} also propose linearly learnt PSD for garments, but conditioned on temporal features processed by an RNN to achieve a non-linear mapping. Later, Santesteban et al. \cite{santesteban2019learning} propose an explicit non-linear mapping for PSD through an MLP for a single template garment. The main drawback of these approaches is that PSD must be learnt for each template garment, which in turns requires new simulations to obtain the corresponding data. To address this issue, many researchers propose an extension of a human body model (SMPL \cite{loper2015smpl}), encoding garments as additional displacements and topology as subsets of vertices \cite{alldieck2018video, alldieck2019tex2shape, bertiche2020cloth3d, bhatnagar2019multi, patel2020tailornet}. Alldieck et al. \cite{alldieck2018video, alldieck2019tex2shape} propose a single model for body and clothes, first as vertex displacements and later as texture displacement maps, to infer 3D shape from single RGB images. Similarly, Bhatnagar et al. \cite{bhatnagar2019multi} also learn a space for body deformations to encode outfits, plus an additional segmentation to separate body and clothes, also to infer 3D garments from RGB. Jiang et al. \cite{jiang2020bcnet} propose 3D outfit retrieval from images and predicting the corresponding blend weights w.r.t. SMPL skeleton using, as labels, the weights of the nearest skin vertices. Patel et al. \cite{patel2020tailornet} encode a few different garment types as subsets of body vertices and propose a strategy to explicitly deal with high frequency pose-dependant cloth details for different body shapes and garment styles. Bertiche et al. \cite{bertiche2020cloth3d} encode thousands of garments on top of the human body by masking its vertices. They learn a continuous space for garment types, on which later they condition, along with the pose, the vertex deformations. Using a body model to represent garments allows handling multiple types with a single model. Nonetheless, it is still limited to single garments, as it cannot work with multiple layers of cloth. For the same reason, they cannot handle complex garment details. This reduces their applicability in real scenarios. Our proposed methodology allows working with arbitrary topologies, number of layers and complex details. Additionally, output format is highly efficient and allows easy integration into graphics engines, increasing compatibility and applicability.

\section{Predicting Animated 3D Models}
\label{sec:theory}

Computer graphics 3D animated models are constructed using skinning and/or blend shapes. In the former, given a 3D mesh with $N$ vertices as $\mathbf{T}\in\mathbb{R}^{N\times3}$ and a skeleton with $K$ joints as $\mathbf{J}\in\mathbb{R}^{K\times3}$, each mesh vertex is attached to each joint with a blend weights matrix $\mathbf{W}\in\mathbb{R}^{N\times K}$. Then, animating the 3D mesh can be achieved by posing skeleton $\mathbf{J}$ through linear transformation matrices (rotation, scaling and translation). Vertex transformation matrices are obtained as a weighted average of joint transformations as described by blend weights. For realistic human and garment animation, only rotations are applied to the joints, and thus, an axis-angle representation is used for pose as $\theta\in\mathbb{R}^{K\times3}$. For the latter, given $\mathbf{T}$ as defined above, a blend shapes matrix as $\mathbf{D}\in\mathbb{R}^{M\times N\times3}$ encodes $M$ different deformations (shapes) $D_i\in\mathbb{R}^{N\times3}$ for $\mathbf{T}$. To animate the mesh, $M$ shape keys are required. These keys are used to linearly combine blend shapes to obtain a final deformation for $\mathbf{T}$. Temporal evolution of shape keys animates the mesh. More complex 3D models use a combination of both techniques. First, $\mathbf{T}$ is linearly deformed through blend shapes and later posed along skeleton $\mathbf{J}$ according to blend weights. Whenever shape keys are defined as a function of skeleton pose, we have Pose Space Deformations driven by \textit{pose} keys. More formally, in human and garment animation domain:
\begin{equation}
    \mathbf{V}_{\theta} = W(\mathbf{T} + \sum_i^M f(\theta)_i D_i, \mathbf{J}, \theta, \mathbf{W}) %\mid d\mathbf{T}_\theta = \sum_i^M f(\theta)_i D_i,
    \label{eq:skinning}
\end{equation}
%\begin{equation}
%    d\mathbf{T}_\theta = \sum_i^M f(\theta)_i D_i
%\end{equation}
Where $W(\cdot)$ is the skinning function that poses mesh vertices as described by $\mathbf{J}$ and $\theta$, $\mathbf{V}_{\theta}$ is the posed vertices, $f(\cdot)$ is a function that maps pose $\theta$ to $M$ pose keys and $D_i$ are the shapes within blend shapes matrix $\mathbf{D}$. These techniques are the standard for 3D animation. All current graphics engines are compatible with these methods.

An example for this is SMPL \cite{loper2015smpl} (human body model). SMPL consists of a template mesh with vertices $\mathbf{T}\in\mathbb{R}^{6890\times3}$, an skeleton $\mathbf{J}\in\mathbb{R}^{24\times3}$, a blend weights matrix $\mathbf{W}\in\mathbb{R}^{6890\times24}$ and two blend shapes matrices, one to represent different body shapes, $\mathbf{D}_{shape}\in\mathbb{R}^{10\times6890\times3}$, and another for Pose Space Deformations, $\mathbf{D}_{PSD}\in\mathbb{R}^{207\times6890\times3}$. Body shape is defined by shape keys $\beta\in\mathbb{R}^{10}$ and Pose Space Deformations by pose keys as flattened rotation matrices (removing global orientation) $R\in\mathbb{R}^{207}$. Because of its formulation SMPL is compatible with current graphics engines. Through this paper, we use SMPL as support body model for animating outfits.

In this work we present a novel approach for garment animation. While recent works are already leveraging skinning blend weights w.r.t. body skeleton to drive garment motion, authors usually rely on complex formulations for Pose Space Deformations, hindering their compatibility with graphics engines and reducing significantly their applicability in real scenarios. We propose learning a mapping from template outfit (canonical pose) meshes to their corresponding blend weights and blend shapes matrices through deep learning. That is, learning a neural network $\mathcal{M}$ as:
\begin{equation}
    \mathcal{M} : \{\mathbf{T},\mathbf{F}\} \rightarrow \{\mathbf{W}, \mathbf{D}_{PSD}\},
    \label{eq:map}
\end{equation}
Where $\mathbf{T}$ are outfit template vertices and $\mathbf{F}$ is mesh faces, $\mathbf{W}$ and $\mathbf{D}_{PSD}$ are the blend weights and blend shapes matrices as defined above. Note that in deployment, a template outfit is processed by the network only once into its standard animated 3D model format. Once blend weights and blend matrices are obtained, the outfit is used as any other 3D animated model. This makes predictions automatically compatible with all graphics engines, and furthermore, due to the exhaustive optimization of rendering pipelines for such models, it is an extremely computationally efficient representation. This further extends its applicability to portable devices and low-computing environments. It represents an advantage against other related works that predict vertex locations directly with neural networks (and often through large, complex models). Such approaches require major engineering efforts to adapt to real applications. Furthermore, due to memory footprint and computational cost of neural networks, these solutions might be impossible to use in low-computing devices. Finally, we also show how this approach allows generalization to unseen template outfits without retraining, which greatly enhances scalability.

\section{Methodology}
\label{sec:method}

%In this section, we formally describe the problem and present the chosen architecture and its components.%for the model along with detailed explanations of each of its components.

Given PBS data for outfits on top of human bodies (SMPL) in different action sequences, we define samples $\mathcal{S} = \{X, Y\}$ as $X = \{\mathbf{T}, \mathbf{F}, \theta, \beta, g\}$ and $Y = \{\mathbf{V}_{PBS}\}$, where $\mathbf{T}$ is the template outfit vertices (canonical pose), $\mathbf{F}$ is outfit mesh faces, $\theta$ is body skeleton pose, $\beta$ is body shape parameters, $g$ is body gender and $\mathbf{V}_{PBS}$ is the outfit vertex locations in the simulated data. Our goal is to train $\mathcal{M}$ as defined in Eq. \ref{eq:map} such that $\mathbf{W}$ and $\mathbf{D}_{PSD}$ yield $\mathbf{V}_{PBS}$ after applying Eq. \ref{eq:skinning} (Note that for SMPL, skeleton is a function of shape $\beta$ and gender).

\subsection{PBS Data and Physical Consistency}
\label{sec:phys_consistency}

The mapping from pose-space to outfit-space is a multi-valued function. Different simulators, initial conditions, action speeds, timesteps and integrators, among other factors, will generate different valid outfit vertex locations for the same body pose and shape and outfit. Training on PBS data falsely assumes that this mapping is single-valued. Samples with similar $X$ but significantly different $Y$ will hinder network performance during training and most likely converge to average vertex location under a supervised loss. Moreover, a final user does not know the \textit{ground truth} and therefore cannot perceive the accuracy of the model, but the user can assess the physical consistency of the predictions (collision-free and cloth consistency). Because of this, while resorting to PBS data for supervision is helpful for training networks, minimizing Euclidean error w.r.t. \textit{ground truth} does not guarantee physical consistency, and therefore, the applicability of the predictions in real life is limited. Recent works \cite{santesteban2019learning, patel2020tailornet} propose post-processing to solve body penetrations. This partially defeats the purpose of using deep-learning and further compromises method compatibility and performance. We propose combining supervised training with unsupervised physically based training to alleviate the need of post-processing.

Physical consistency is a crucial part of proper outfit animation. While other approaches develop complex solutions to better overfit to their PBS data and translate training wrinkles to predictions, their lack of physical constraints is detrimental for their usability in real applications. Physical consistency is not only limited to collisions, but also to edge distortion and surface quality. Abnormally stretched or compressed edges (w.r.t. to its template lengths) will create texture distortions (UV map edges do not change in length, but mesh edges do). Approaches that represent garments as a subset of body vertices cannot enforce an edge constraint, as their template is the body itself (original template is lost after registration against the body). Our proposal addresses garment animation independently of edition/resizing, and therefore, it is possible to leverage the original template outfits to enforce the edge constraint during learning. %Fig. \ref{fig:sota_phys} shows qualitative samples of physic-related issues. Samples are gathered from TailorNet \cite{patel2020tailornet} and CLOTH3D \cite{bertiche2020cloth3d} papers. Both approaches rely on body model for garment representation. On the textured sample (TailorNet) we observe an evident pattern distortion. In the middle sample (TailorNet) we see body-to-cloth collision (after post-processing). Also, since these approaches cannot represent more than one garment at a time, to predict outfits, individual garment models are run, which are prone to cloth-to-cloth collisions, as can be seen. Finally, rightmost samples (CLOTH3D) show very noisy garment boundaries (before post-processing).

\subsection{Architecture}

\begin{figure}[t]
    \centering
    \includegraphics[width=1\columnwidth]{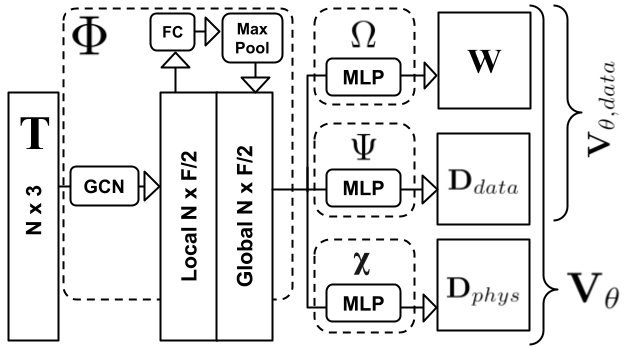}
    \caption{Model overview. The input of the model is a template outfit mesh (with no fixed topology, vertex order or connectivity). We apply graph convolutions to obtain vertex local descriptors. Then, local descriptors are processed by a fully-connected layer and aggregated through per-outfit max-pooling. This yields a global outfit descriptor that is concatenated with each vertex local descriptor. Final vertex descriptors are processed through different MLPs to obtain blend weights $\mathbf{W}$ and blend shapes matrices $\mathbf{D}_{data}$ and $\mathbf{D}_{phys}$. Blend shapes matrices are combined into $\mathbf{D}_{PSD}$, which is used as described in Eq. \ref{eq:skinning} to obtain final predictions. Pose keys for blend shapes matrix are obtained by passing $\theta$ through an MLP with $4$ layers (not shown). \vspace{-.35cm}}
    \label{fig:pipeline}
\end{figure}

The chosen architecture needs to be able to: a) handle unstructured meshes (no fixed vertex order or connectivity) and b) compute non-linear deformations w.r.t. the pose $\theta$ (as cloth behaviour is highly non-linear). To do so, we define the following components: $\Phi : \mathbb{R}^{N\times3} \rightarrow \mathbb{R}^{N\times F}$, $\Omega : \mathbb{R}^{N\times F} \rightarrow \mathbb{R}^{N\times K}$, $\Psi : \mathbb{R}^{N\times F} \rightarrow \mathbb{R}^{P\times N\times3}$ and $\chi : \mathbb{R}^{N\times F} \rightarrow \mathbb{R}^{P\times N\times3}$.
%\begin{equation}
%    \Phi : \mathbb{R}^{N\times3} \rightarrow \mathbb{R}^{N\times F}
%\end{equation}
%\begin{equation}
%    \Omega : \mathbb{R}^{N\times F} \rightarrow \mathbb{R}^{N\times K}
%\end{equation}
%\begin{equation}
%    \Psi : \mathbb{R}^{N\times F} \rightarrow \mathbb{R}^{P\times N\times3}
%\end{equation}
%\begin{equation}
%    \chi : \mathbb{R}^{N\times F} \rightarrow \mathbb{R}^{P\times N\times3}
%\end{equation}
Component $\Phi$ computes per-vertex high-level $F$-dimensional descriptors with local and global information from template outfit mesh (with $F = 512$), $\Omega$ computes per-vertex blend weights from vertex descriptors, $\Psi$ generates a blend shapes matrix supervisedly (note that it is equivalent to per-vertex \textit{blend shapes} matrices as $\mathbf{d}\in\mathbb{R}^{P\times3}$) and $\chi$ generates a blend shapes matrix unsupervisedly that will yield physical consistency. Note that we define $P$ pose keys for blend shapes matrices, instead of the dimensionality of pose $\theta$. We pass $\theta$ through an MLP to obtain a high-level embedding of pose as $\Theta\in\mathbb{R}^P$. The motivation for this is: a) controlling dimensionality $P$, and therefore, blend shapes matrix size and capacity and b) to allow modelling non-linearities from pose-space to vertex-space.

Fig. \ref{fig:pipeline} shows an overview of the model. To learn $\Phi$, we use $4$ layers of graph convolutions applied to template mesh. This will yield a local descriptor, with no global information. Inspired by PointNet \cite{qi2017pointnet}, we process each local descriptor through an additional fully-connected layer and aggregate all vertex descriptors through max-pooling (per outfit). We concatenate this global descriptor to each vertex local descriptor. Then, $\Omega$, $\Psi$ and $\chi$ are defined as MLPs, with $4$ fully-connected layers each, applied to vertex descriptors (vertices are independent \textit{samples}). The chosen architecture permits processing unstructured meshes with any vertex number, order and connectivity. This is a significant advantage against approaches that rely on the body model for garment representation \cite{bertiche2020cloth3d, patel2020tailornet}, since it requires an expensive registration for each sample that introduces error in the data. %Predicted blend weights need to be normalized per-vertex as $w_{ik} = w_i / \sum_k^Kw_ik$ to fulfill the required animation standard (vertex transformations are the weighted average of joint transformations). 
Then, $\Psi$ and $\chi$ both compute blend shapes matrices: $\mathbf{D}_{data}$ to minimize supervision loss and $\mathbf{D}_{phys}$ for physical consistency. Despite being independent branches, on deployment, both matrices are combined to obtain the final PSD matrix $\mathbf{D}_{PSD} = \mathbf{D}_{data} + \mathbf{D}_{phys}$, thus keeping the aforementioned compatibility with graphics engines. Finally, the MLP used to obtain the high-level pose embedding $\Theta$ consists on $4$ fully-connected layers. The output of the model during training is $\mathbf{V}_{\theta,data}$ for $\mathbf{D}_{data}$ and $\mathbf{V}_{\theta}$ for $\mathbf{D}_{PSD}$.

\subsection{Training}

Our model combines both supervised and unsupervised training. The supervised part of the model corresponds to $\Phi$, $\Omega$ and $\Psi$. The goal of this \textit{submodel} is to minimize Euclidean error w.r.t. PBS data. Thus, for its training, we apply an standard L2 loss on predicted vertex locations:
\begin{equation}
    \mathcal{L}_{data} = \sum\|\mathbf{V}_{\theta,data} - \mathbf{V}_{PBS}\|^2,
\end{equation}
Then, the unsupervised part of the model corresponds only to $\chi$. We define unsupervised losses to satisfy prior distributions based on physical constraints. First, to ensure cloth consistency of predictions, inspired by \textit{mass-spring} model (most widely used PBS model for cloth), we define a cloth loss term as:
\begin{equation}
    \mathcal{L}_{cloth} = \mathcal{L}_{E} + \lambda_B\mathcal{L}_{B} = \sum_{e\in E}\| e - e_{\mathbf{T}}\|^2 + \lambda_B\Delta(\mathbf{n})^2,
\end{equation}
where $\mathcal{L}_E$ is the edge term and $\mathcal{L}_B$ is the bending term. Then, $E$ is the set of edges of the given outfit mesh, $e$ is the predicted edge length and $e_{\mathbf{T}}$ is the edge length on the template outfit $\mathbf{T}$. Then, $\Delta(\cdot)$ is the Laplace-Beltrami operator applied to vertex normals $\mathbf{n}$ of the predicted outfit and $\lambda_B$ balances both losses. $\mathcal{L}_{E}$ enforces the output meshes to have the same edge lengths as the input template outfit, while $\mathcal{L}_{B}$ helps yielding locally smooth surfaces, as it penalizes differences on neighbouring vertex normals. To avoid excessive flattening, we choose $\lambda_B = 0.0005$. Then, to handle collisions against the body, we define a loss as:
\begin{equation}
    \mathcal{L}_{collision} = \sum_{(i,j)\in A} min(\mathbf{d}_{j,i}\cdot \mathbf{n}_j - \epsilon, 0)^2,
    \label{eq:collision}
\end{equation}
where $A$ is the set of correspondences $(i,j)$ between predicted outfit and body through nearest neighbour, $\mathbf{d}_{j,i}$ is the vector going from the $j$-th vertex of the body to the $i$-th vertex of the outfit, $\mathbf{n}_j$ is the $j$-th vertex normal of the body and $\epsilon$ is a small positive threshold to increase robustness. This loss is a simplified formulation that assumes cloth is close to the skin, and penalizes outfit vertices placed inside the skin. In our experiments, we choose $\epsilon = 5$mm. Thus, the unsupervised loss is defined as:
\begin{equation}
    \mathcal{L}_{phys} = \mathcal{L}_{cloth} + \lambda_{collision}\mathcal{L}_{collision}
\end{equation}
where $\lambda_{collision}$ is the balancing weight for the collision term (around $2$-$10$ in our experiments). Note how both terms $\mathcal{L}_{cloth}$ and $\mathcal{L}_{collision}$ are defined as priors (based only on $X$, not on $Y$). We define an additional loss term as an L2 regularization on deformations due to $\chi$ with a balancing weight $\lambda = 1e-2$. This leads $\chi$ to use deformations as small as possible to solve physical constraints. While the whole model is differentiable and could be trained end-to-end, we back-propagate $\mathcal{L}_{phys}$ only through $\chi$. The motivation for this is:
\begin{itemize}[noitemsep,topsep=0pt,parsep=0pt,partopsep=0pt]
    \item \textbf{Independent Tasks.} We empirically observed how supervised and unsupervised terms fight each other, compromising one or both tasks. Thus, by training different parts of the model independently, we do not need to find a balance between low Euclidean error and physical consistency. This allows the supervised submodel to learn the main deformations leveraging PBS data and the unsupervised branch to enforce physical consistency without their gradients hindering each other.
    \item \textbf{Unsupervised Training.} Since $\mathcal{L}_{phys}$ does not rely on $Y$, it is possible to train $\chi$ with new samples where $\theta$ is replaced in $X$ by any other sample pose. This increases the amount of available data to train, enhancing generalization of physical consistency.
\end{itemize}
In practice, it is not helpful to train $\chi$ until the supervised training has converged.

\section{Experiments}
\label{sec:ablation}

From the public datasets on garments, only CLOTH3D \cite{bertiche2020cloth3d} contains enough outfit variability to implement this approach and achieve proper generalization. It contains $\sim7.5$k sequences, each with a different template outfit in rest pose plus up to $300$ frames. The outfits are simulated on top of an animated 3D human (SMPL), each with a different body shape. Likewise, we use SMPL skeleton in Eq. \ref{eq:skinning}, so it drives the motion of the outfit, and its body mesh in Eq. \ref{eq:collision}. For the ablation study, we subsample $50$k training \textit{frames} and $5$k test \textit{frames} from CLOTH3D in a stratified manner w.r.t. sequences without outfit overlapping between both sets. Each model is trained for $10$ epochs. We additionally present proof-of-concept computer vision applications as well as a performance analysis in the supplementary material.

\subsection{Ablation study}

\begin{table}[]
\centering
\begin{tabular}{lc}
\multicolumn{1}{c}{} & Euclidean error (mm) \\ \hline
Baseline & 29.98 \\
+Global & 28.04 \\
+GlobalLite & 28.59 \\
+Global+GCN & 28.76 \\
+Global with MLP & 28.43 \\
DeePSD & \textbf{25.13} \\ 
-without pose embedding & 30.93 \\ \hline
\end{tabular}
\caption{Architecture ablation study. First, as a baseline, we train $\Omega$ and $\Psi$ to predict vertex deformations instead of blend shapes matrices. Subsequent rows are baselines extensions (deformation prediction) with a global descriptor. \textit{DeePSD} row corresponds to the architecture shown in Fig. \ref{fig:pipeline}. As it can be seen, predicting blend shapes matrices is the best performing approach.}
\label{tab:architecture}
\end{table}

\begin{table}[]
\centering
\begin{tabular}{lc}
\multicolumn{1}{c}{} & Euclidean error (mm) \\ \hline
DeePSD & 25.13 \\
+ SMPL shape/gender & 25.15 \\
+ Fabric & 24.76 \\
+ Tightness + Fabric & \textbf{24.66} \\
+ SMPL + Tightness + Fabric & 25.01 \\ \hline
\end{tabular}
\caption{Conditioning to metadata available in CLOTH3D \cite{bertiche2020cloth3d} for each sample. We concatenate metadata to each vertex descriptor: SMPL shape and gender, per-garment fabric and per-outfit tightness. As shown, body metadata hinders performance, while outfit metadata enhances it.}
\label{tab:metadata}
\end{table}

\begin{table}[]
\centering
\begin{tabular}{lcccc}
\multicolumn{1}{c}{} & Error & \multicolumn{1}{l}{Edge} & \multicolumn{1}{l}{Bend} & \multicolumn{1}{l}{Collision} \\ \hline
No phys. & 24.66 & 1.27 & 0.031 & 11.59\% \\
Phys. & 33.75 & 1.13 & 0.029 & 1.29\% \\
+poses & 34.45 & 1.12 & 0.029 & 1.02\%  \\ \hline
\end{tabular}
\caption{Unsupervised training. We measure cloth quality with average edge elongation/compression and bending angle between neighbouring vertex normals. For body collision, we show the ratio of vertices placed within the body.}
\label{tab:chi}
\end{table}

%\begin{table}[]
%\centering
%\begin{tabular}{lcl}
%\multicolumn{1}{c}{} & Outfit & Poses \\ \hline
%Tshirt & 26.05 & \multicolumn{1}{c}{26.48} \\
%Top & 18.59 & \multicolumn{1}{c}{19.85} \\
%Trousers & 15.46 & \multicolumn{1}{c}{15.47} \\
%Jumpsuit & 18.51 & \multicolumn{1}{c}{17.96} \\
%Skirt & \multicolumn{1}{l}{43.65} & 38.71 \\
%Dress & \multicolumn{1}{l}{37.19} & 35.08 \\ \hline
%Total (outfit-wise) & \multicolumn{1}{l}{25.83} & 24.82 \\ \hline
%\end{tabular}
%\end{table}

\begin{table}[]
\centering
\begin{tabular}{lc}
\multicolumn{1}{c}{} & Euclidean error (mm) \\ \hline
Tshirt & 25.77 \\
Top & 17.33 \\
Trousers & 14.50 \\
Jumpsuit & 17.23 \\
Skirt & 41.15 \\
Dress & 35.94 \\ \hline
Total & 23.95 \\ \hline
\end{tabular}
\caption{Final quantitative results per garment. Note how \textit{tighter} garment types have a significantly lower error than others.}
\label{tab:final}
\end{table}

\begin{figure*}
    \centering
    \includegraphics[width=.85\textwidth]{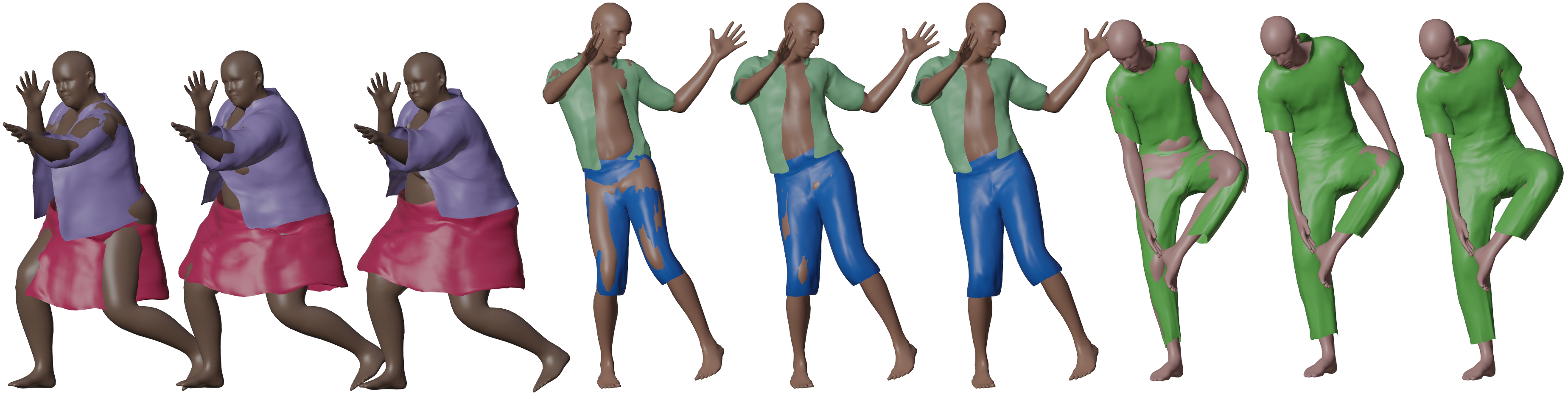}
    \vspace{-0.3cm}
    \caption{Qualitative results obtained by enforcing physical consistency. For each sample we show the results of each experiment in Tab. \ref{tab:chi} in the same order from left to right.     \vspace{-0.5cm}}
    \label{fig:chi}
\end{figure*}

We first evaluate the supervised part of the model ($\Phi$, $\Omega$ and $\Psi$) by using the average vertex Euclidean error per outfit. In Tab. \ref{tab:architecture} we show the results to justify the design of the network. First, we propose a baseline model. In this baseline, global descriptor is not computed and $\Psi$ predicts vertex deformations instead of blend shapes matrices by concatenating pose to vertex descriptors. The following models are modifications of the baseline (predict deformations). The second row shows the result obtained by using global descriptors. It improves the accuracy of the predictions. The third row corresponds to a model with a lower descriptor dimensionality ($F = 128$), and we observe an slight increase in error. In the next experiment, we implement $\Omega$ and $\Psi$ as graph convolutions instead of fully-connected layers. This worsens results at the cost of extra computational cost, thus we discard the use of graph convolutions after $\Phi$. Note that this behaviour is expected, as global descriptor is broadcasted through vertices, and therefore, convolutions perform redundant information passes that hinder the learning. The next row corresponds to a model where global descriptor is obtained by replacing the single fully-connected layer in $\Phi$ with a MLP. Performance does not improve. \textit{DeePSD} row corresponds to the architecture shown in Fig. \ref{fig:pipeline}. As one can observe, predicting blend shapes matrices instead of vertex deformations not only increases model compatibility with graphics engines, but it also improves performance. The final row corresponds to the same architecture as DeePSD, but using pose $\theta$ as pose keys instead of a high-level pose embedding. We see that predictions are less accurate, thus pose embedding $\Theta$ is beneficial.

We consider the effect of including additional metadata present in CLOTH3D. That is, SMPL body shape and gender, garment-wise fabric labels and outfit-wise tightness values. We combine these metadata by concatenating them to each vertex descriptor. Tab. \ref{tab:metadata} shows the quantitative results. The first row corresponds to the best model of Tab. \ref{tab:architecture}. Each next row is named after the metadata used. As it can be observed, outfit metadata reduces Euclidean error while body metadata appears to be detrimental.

To evaluate the unsupervised model, we design suitable metrics for assessing cloth quality and physical constraints:
\begin{itemize}[noitemsep,topsep=0pt,parsep=0pt,partopsep=0pt]
    \item \textbf{Edge Length.} Length difference between predicted and rest outfit edges, expressed in millimeters.
    \item \textbf{Bend Angle.} Cosine distance for pairs of neighbouring vertex normals.
    \item \textbf{Collision.} Ratio of collided vertices.
\end{itemize}
Edge metric summarizes cloth integrity. Cloth needs to compress or stretch to fit its environment in real life and PBS, thus, a zero-valued edge error might be impossible (even undesirable). Nonetheless, an abnormally high value suggests distorted predictions. Similarly, bend angle cannot be zero, otherwise we have a completely flat surface. Again, high values for this metric show poor cloth quality. Finally, for collisions, a zero-valued metric means physically consistent predictions. In practice, the training data contains invalid combinations of pose and shape (bodies with self-collisions), and therefore, a $0\%$ of collided vertices is impossible. Tab. \ref{tab:chi} shows the results for the ablation study of the physical consistency. First, we evaluate the predictions obtained with supervised loss only (best model of Tab. \ref{tab:metadata}). Second row shows the results obtained with $\chi$ trained without pose augmentation. The third row shows the results after training each sample with randomly chosen poses. We can observe that while Euclidean error increases, physical related metrics improve, specially collision. The model is learning to predict outfits farther from \textit{ground truth} PBS data, but with higher physical consistency. As explained in Sec. \ref{sec:phys_consistency}, physical consistency cannot be summarized in one or few quantitative metrics. Results must be evaluated qualitatively. Fig. \ref{fig:chi} shows a qualitative comparison of these experiments. As it can be seen, without physical constraints, although predictions have lower error by a large margin, qualitatively they are much worse. Also, we see that training unsupervisedly with randomly sampled poses further improves generalization.

We report final supervised results after fine-tuning with all data on Tab. \ref{tab:final}. We decompose the error per garment. Note that \textit{T-shirt} includes open shirts as well. We observe a worse performance for skirts and dresses. We also find a high error on \textit{T-shirt}, likely due to open shirts. This is an expected behaviour, since modelling garments statically through skinning assumes the cloth will follow the body motion. Loose garments show a much more complex dynamics, and thus, static approaches will fail to model such garments. Fig. \ref{fig:title} shows qualitative results. We can see how the model can generalize to unseen complex outfits without retraining. Additionally, while cloth-to-cloth interaction is not explicitly addressed, the model is able to deal with multiple layers of cloth. It shows it can also handle complex geometric details (chest flower). As stated, it maintains cloth consistency, thus no artifacts appear on texturing. Finally, due to the unsupervised blend weights learning, skirts are robust against skinning artifacts due to leg motion (see supplementary material for more details on blend weights).

\subsection{Comparison with related works}

\begin{table}[]
\centering
\begin{tabular}{lc}
\multicolumn{1}{c}{} & Euclidean error (mm) \\ \hline
CLOTH3D \cite{bertiche2020cloth3d} & 29.0 \\
DeePSD & \textbf{23.78} \\ \hline
\end{tabular}
\caption{Comparison against CLOTH3D baseline. As CLOTH3D \cite{bertiche2020cloth3d}, we report the error garment-wise.\vspace{-.5cm}}
\label{tab:sota}
\end{table}

\begin{figure}[!t]
    \centering
    \vspace{-.3cm}
    \includegraphics[width=.85\columnwidth]{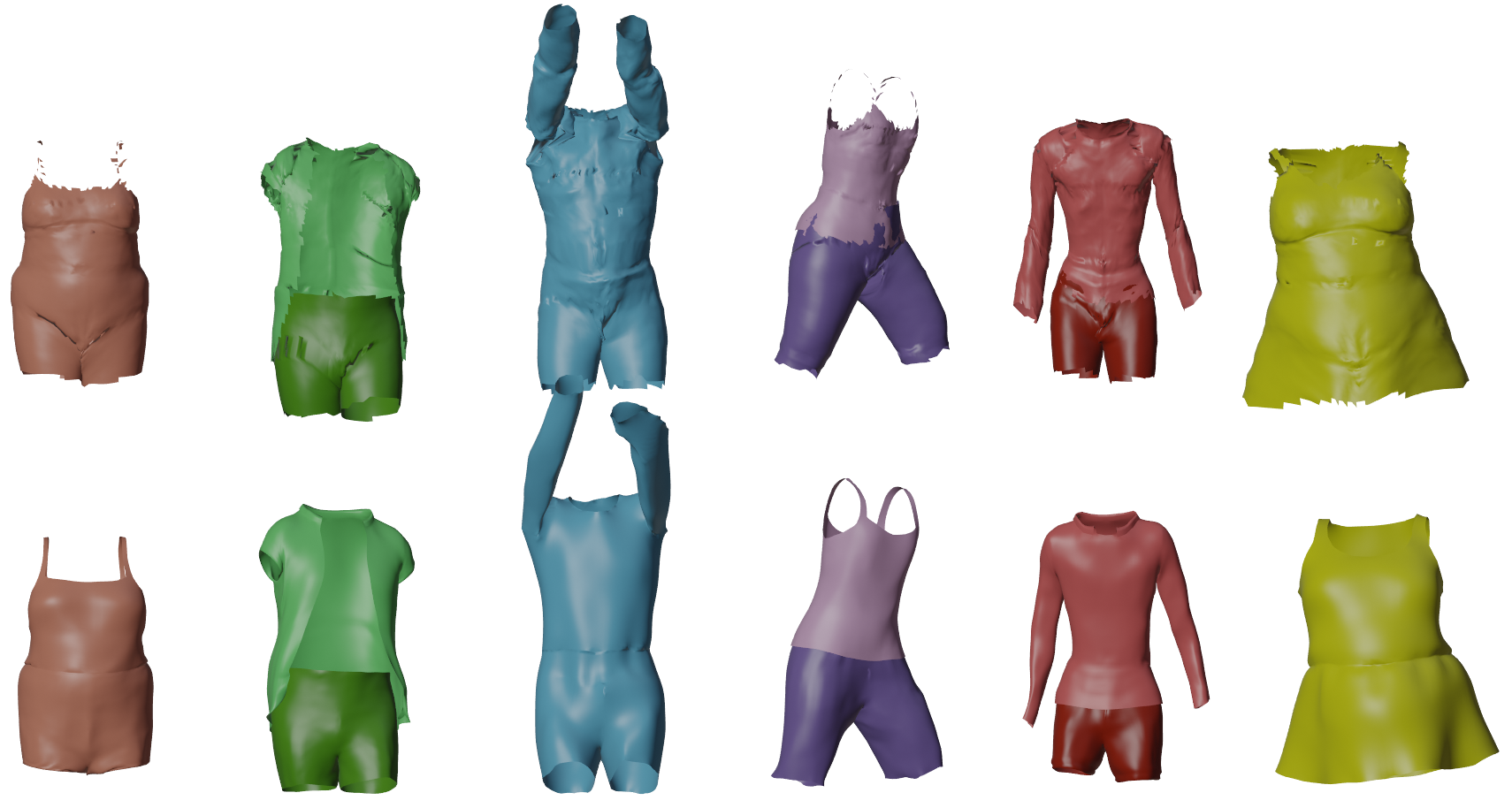}
    \vspace{-.3cm}
    \caption{Qualitative comparison against CLOTH3D \cite{bertiche2020cloth3d} baseline. Upper row: CLOTH3D. Lower row: DeePSD. \vspace{-.4cm}}
    \label{fig:vs_cloth3d}
\end{figure}

%\begin{figure}
%    \centering
%    \includegraphics[width=.8\columnwidth]{img/tailornet_paper_imgs.png}
%    \caption{Figures extracted from TailorNet \cite{patel2020tailornet}. Texture shows distortion due to the lack of edge constraint. We also see how post-processing could not completely solve collision between skirt and body. Cloth-to-cloth collisions appear as well. In the lower row, we show samples with different pose, body shape and/or garment style. Note how wrinkles around neck and shoulders are very similar.}
%    \label{fig:tailornet_paper}
%\end{figure}

\begin{figure}
    \centering
    \vspace{-.3cm}
    \includegraphics[width=\columnwidth]{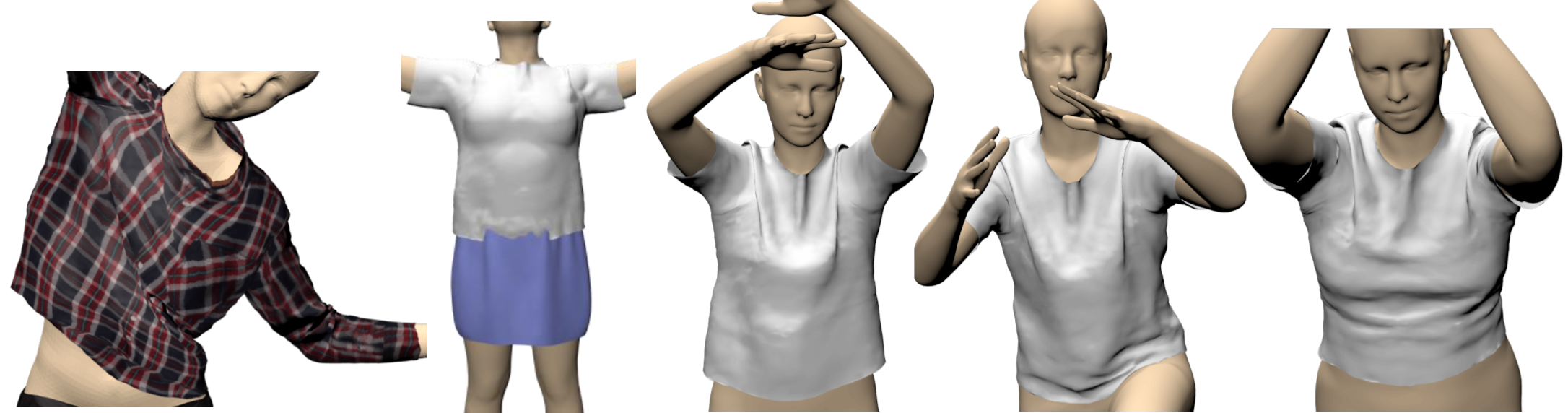}
    \caption{Figures extracted from TailorNet \cite{patel2020tailornet}. Texture shows distortion due to the lack of cloth consistency. We see cloth-to-cloth collisions appear. Next, we show samples with different pose, body shape and/or garment style. Note how wrinkles around neck and shoulders are very similar.}
    \label{fig:tailornet_paper}
\end{figure}

\begin{figure*}
    \centering
    \includegraphics[width=.85\textwidth]{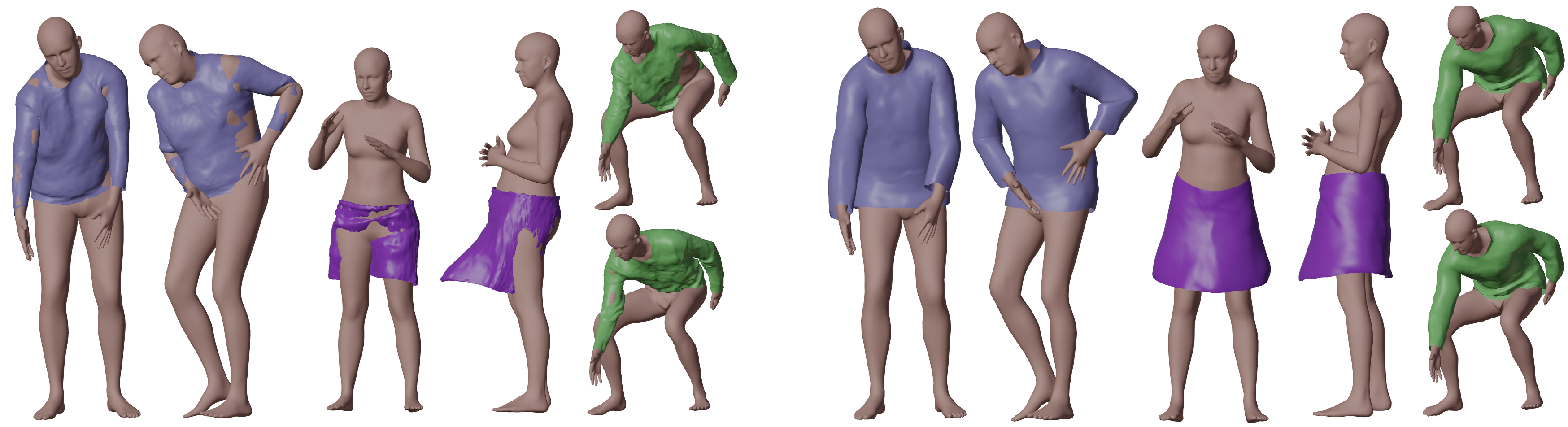}
    \vspace{-0.3cm}
    \caption{Comparison with TailorNet. Left: TailorNet. Right: DeePSD. TailorNet heavily relies in post-processing for valid predictions and generates noisy surfaces. The third sample (green T-shirt) shows two consecutive frames, note how TailorNet cannot guarantee temporal consistency.    \vspace{-0.4cm}}
    \label{fig:tailornet0}
\end{figure*}

\textbf{CLOTH3D.} We compare DeePSD with CLOTH3D baseline quantitatively in Tab. \ref{tab:sota} and qualitatively in Fig. \ref{fig:vs_cloth3d}. As it can be seen, our method outperforms CLOTH3D baseline. On the one hand, CLOTH3D baseline shows noisy boundaries and even broken suspenders. Furthermore, we observe the body geometry is present in the CLOTH3D reconstructed garment due to the use of SMPL body for garment representation. On the other hand, since DeePSD uses the original templates, boundaries are smooth and there is no bias to body geometry. Additionally, in spite of not dealing directly with cloth-to-cloth collisions, it appears that DeePSD is more robust in this aspect.% Nonetheless, note that CLOTH3D model is proposed as a baseline able to deal with garment edition/resizing as well as animation.

\textbf{TailorNet.} A fair quantitative comparison against the work of \cite{patel2020tailornet} is not possible. On one hand, TailorNet original simulations are not public, only the registered version against SMPL body. This means: a) original templates are lost and recovering them for each shape-style pair is unfeasible and b) their dataset has a fixed vertex order and connectivity (SMPL body). Since our main contribution is the generalization to unstructured meshes, comparing our methodology using a dataset with fixed vertex order against a methodology designed specifically for these data cannot be done fairly. On the the other hand, TailorNet is an ensemble of around $20$ MLPs per each garment and gender which makes adapting it to CLOTH3D unfeasible, due to a much higher garment style variability. Thus, we perform a qualitative comparison. In Fig. \ref{fig:tailornet_paper} (extracted from \cite{patel2020tailornet}) we analyze their qualitative results. Note that TailorNet uses post-processing. The first observation is a clear texture distortion due to the lack of cloth consistency. Next, we see cloth-to-cloth collision. TailorNet requires a model for each individual garment, thus, as CLOTH3D baseline, it is prone to these issues. Finally, we visualize three samples with different pose, body shape and/or garment style. All show similar wrinkles, which suggests overfitting to training data. In Fig. \ref{fig:tailornet0} we qualitatively compare TailorNet (left) with DeePSD (right). For fairness, since our approach uses no post-processing, we remove TailorNet post-processing. We gather similar garments and body shapes in TailorNet data and CLOTH3D and compute the same sequences using both models. As it can be seen, TailorNet is highly dependant on its post-processing due to a high amount of collided vertices. For the green T-shirt, samples correspond to consecutive frames. TailorNet cannot keep temporal consistency. DeePSD does not suffer from such effect. Similar to CLOTH3D baseline, we observe how body geometry is present on TailorNet predictions (leftmost sample chest) due to the use of SMPL to represent garments. %Note that TailorNet also handle garment edition/resizing.

TailorNet succeeds in generating wrinkles in their predictions by overfitting an ensemble of MLPs per each garment type and gender. As stated by its authors: \textit{"Our key simplifying assumption is that two garments on two different people will deform similarly..."}.
%This further reinforces our motivation explained in Sec. \ref{sec:phys_consistency}. As long as predictions look like cloth, we can consider them valid. It is well-known that in fact, wrinkles are chaotic and unpredictable (the same person, with the same outfit and the same pose could present totally different wrinkles).
Nonetheless, this has drawbacks. On one hand, as we have seen, it strongly compromises physical consistency, and thus, relies on post-processing. This increases sample generation times by $150$-$300$ms. Note that applying a post-processing eliminates differentiability. Another drawback is the complexity of their model. Their ensemble of MLPs takes around $2$GB per garment and gender. %Thus, TailorNet wrinkle generation capability does not justify its huge complexity and compromise of physical constraints. 
All of this hurts its applicability, compatibility and performance (and then, portability). On the contrary, DeePSD is a single small-sized model ($4.4$MB) that allows animating any outfit (not only individual garments as body homotopies) without retraining. Predictions are generated as highly computationally efficient  models (blend weights and blend shapes) compatible with any graphics engine. We obtain running times of $3$-$6$ms for individual samples and around $0.1$ms for batched samples (depending on vertex count). Furthermore, through physically based unsupervised learning, we alleviate the need of post-processing, thus maintaining differentiability and the aforementioned computational performance.

%\subsection{Performance}

%We train our model in a GTX1080Ti for $10$ epochs, this takes around $8$-$16$ hours  (unsupervised losses are computationally expensive). Our model has a size of $4.4$Mb. For deployment, we run the model once per outfit to obtain animated 3D models in standard format. This format is highly computationally efficient due to exhaustive optimization of current graphics engines and GPUs. The only extra computational cost is obtaining the high-level pose embedding. Since an MLP is the most basic deep-learning model, it is unlikely to find a more efficient deep-based approach for 3D animation. In practice, we get $3$-$6$ms per sample and around $0.1$ms for batched samples, depending on vertex count.

\section{Conclusions and Future Work}
\label{sec:conclusion}

We presented a novel approach for garment animation. Breaking the trend of previous approaches that try to predict vertex deformations through deep learning, we proposed learning a mapping from outfit space to animated 3D model space. We showed how this allows generalization to unseen outfits as well as compatibility with graphics engines. We observed how recent works need to leverage the body model for garment representation to allow edition/resizing along with animation, leading to overly complex models with scalability, compatibility and applicability issues. We addressed these issues by identifying garment animation as an independent task. Also, we prioritized physical consistency in our predictions, thus relieving the need of post-processing. In summary, we developed an efficient approach applicable in real-scenarios as it is, even portable devices, that allows a more intuitive workflow for CGI artists that does not require expert knowledge in deep learning.

We observed limitations in our approach. First, loose garments, such as skirts and dresses, cannot be properly modelled with static approaches. To this end, we set as future work adapting our methodology to work with the temporal dimension. To keep its compatibility, pose keys should be computed with a temporal neural network while the training enforces dynamic learning (whether it is from data or unsupervisedly through physical consistency). We also observed how recent works grow on complexity to model fine geometric details (wrinkles). We believe the best approach to deal with garment wrinkles is through normal map generation because: a) it allows using lower vertex counts without compromising details, b) it is directly compatible with all graphics engines and c) it is more robust to collisions, since graphics engines compute face visibility on base geometry. Current works on this domain appear to be promising \cite{lahner2018deepwrinkles, zhang2020deep}. We set this as future work.

\textbf{Acknowledgements.} This work has been partially supported by the Spanish project PID2019-105093GB-I00 (MINECO/FEDER, UE) and CERCA Programme/Generalitat de Catalunya.) This work is partially supported by ICREA under the ICREA Academia programme and Amazon Research Awards.

\clearpage
{\small
\bibliographystyle{ieee_fullname}
\bibliography{egbib}
}

\section{Computer Vision Applications}

\begin{figure}
    \centering
    \includegraphics[width=\columnwidth]{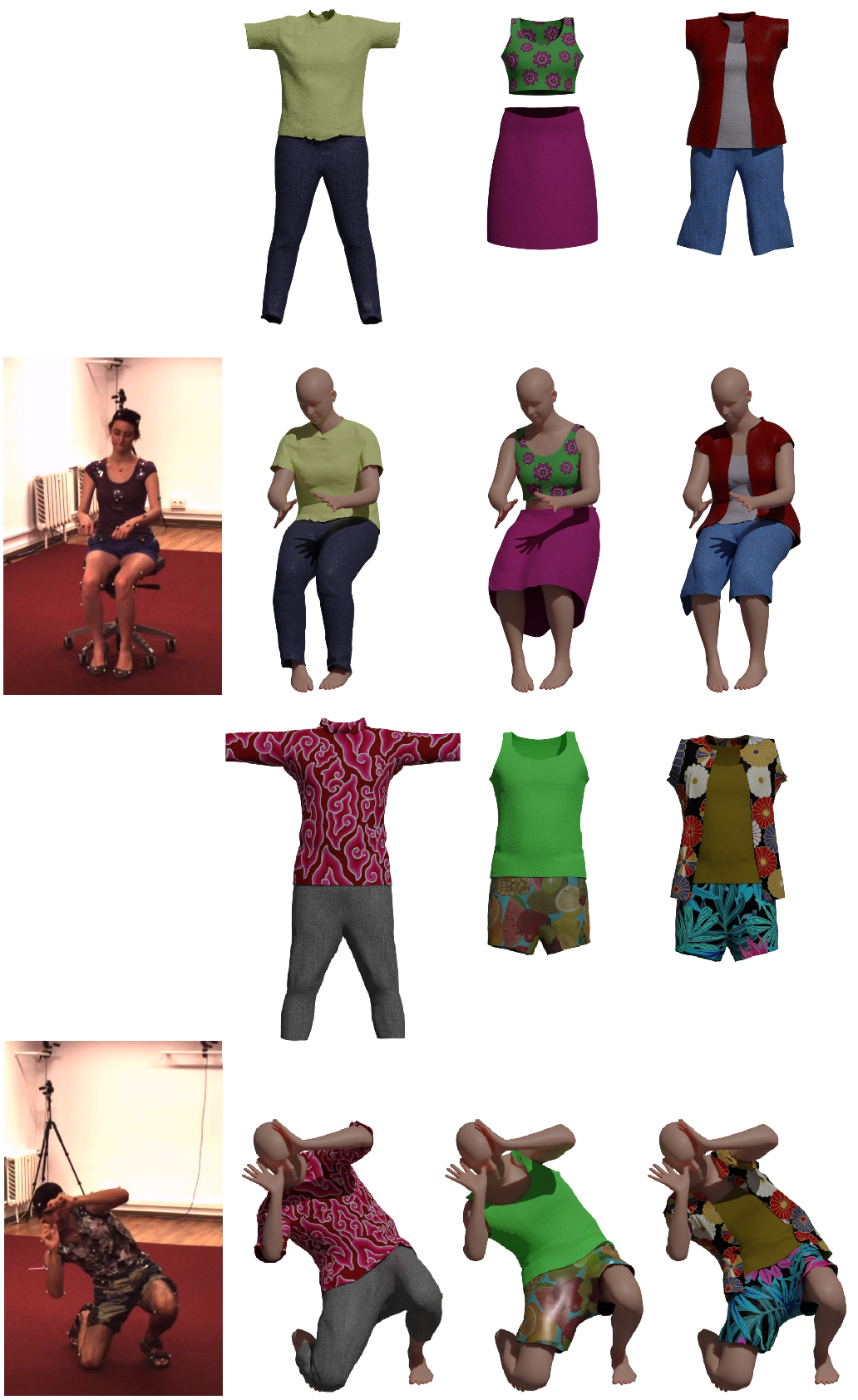}
    \caption{Virtual try-on. Combining DeePSD with a body pose and shape recovery CNN we obtain an effectively working virtual try-on. With the obtained SMPL parameters and a digital wardrobe (3D outfits in canonical pose) it is possible to generate draped 3D models in the corresponding pose. Images extracted from Human3.6M\cite{h36m_pami, IonescuSminchisescu11}}
    \label{fig:try_on}
\end{figure}

\begin{figure}
    \centering
    \includegraphics[width=\columnwidth]{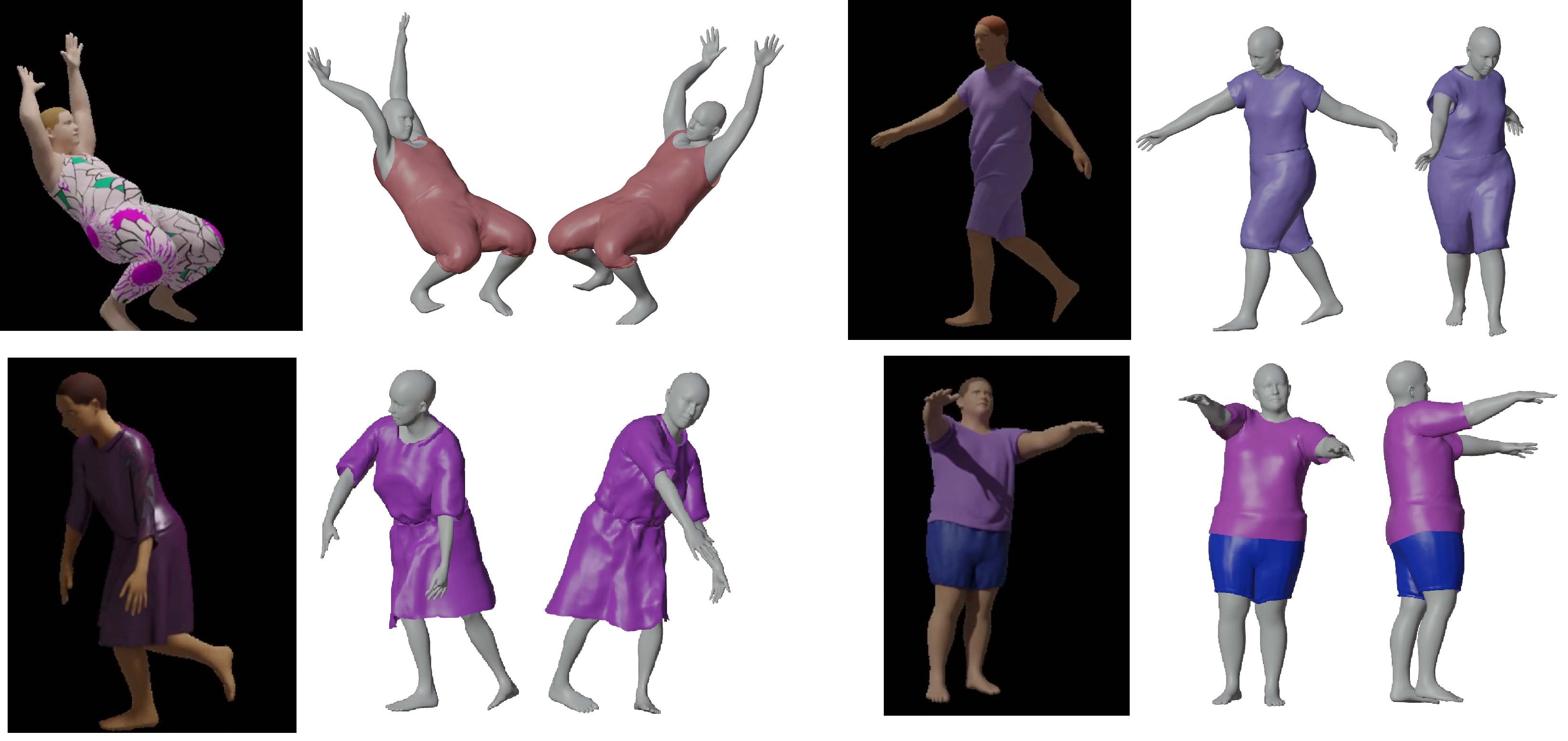}
    \caption{Proof-of-concept computer vision application. We combine an outfit retrieval approach with SMPL parameter regression to obtain 3D predictions of draped humans.}
    \label{fig:cv_app}
\end{figure}

We believe one of the strongest real applications for \textit{DeePSD} is virtual try-ons. Demand for such application is already present and companies in the private sector are pushing towards this direction. We propose a proof-of-concept application. We use an off-the-shelf CNN to regress SMPL parameters from images~\cite{madadi2020smplr}. SMPL parameter predictions are combined with template outfits to generate a 3D model of the subject with the desired outfit. Fig.\ref{fig:try_on} shows the results of this experiment (note that some of these outfits do not appear in CLOTH3D dataset). In practice, for virtual try-ons, the methodology to use has to be scalable to unseen and arbitrary outfits. Thus, methodologies limited to individual garments encoded as body homotopies are not suitable~\cite{bertiche2020cloth3d, patel2020tailornet}. Additionally, we observed how these approaches suffer from texturing artifacts, which further compromises their applicability. Finally, to bring this technology to everyone (portable devices), a model with low computational complexity and memory footprint is necessary. As discussed in the main paper, \textit{DeePSD} has a size of only $4.4$MB and generates animated 3D models, which are extremely efficient and compatible with all graphics engines.

We also present an additional proof-of-concept computer vision application for 3D draped human regression from video using \textit{DeePSD}. Similar to the virtual try-on application, we use a CNN\cite{madadi2020smplr} to regress SMPL pose. Then, to obtain the outfit from the video sequence, we propose the following approach. First, we use \textit{DeePSD} to compute outfit global descriptors (as defined in the main paper) for each outfit in the training set. Then, we train a VGG-16 to regress outfit global descriptors from static frames. Next, in test, we obtain an estimation of the outfit global descriptor for each frame of a given sequence. We retrieve the $5$ nearest neighbours from the training set based on global descriptor Euclidean distance for each frame. We retrieve the outfit with the most appearances. In case of a tie, we choose the one with the lowest Euclidean distance. Finally, once an outfit is retrieved, we combine it with the estimated SMPL pose using \textit{DeePSD} to obtain a final 3D draped human prediction. Fig.\ref{fig:cv_app} shows some samples obtained using this approach. Note that, since template outfit is aligned with the body in canonical pose, we retrieve SMPL shape parameters and gender along with the outfit. As it can be seen, these predictions benefit from the physical consistency of \textit{DeePSD} predictions. 

\section{Blend Weights}
\label{sec:blendweights}

\begin{figure*}
    \centering
    \includegraphics[width=\textwidth]{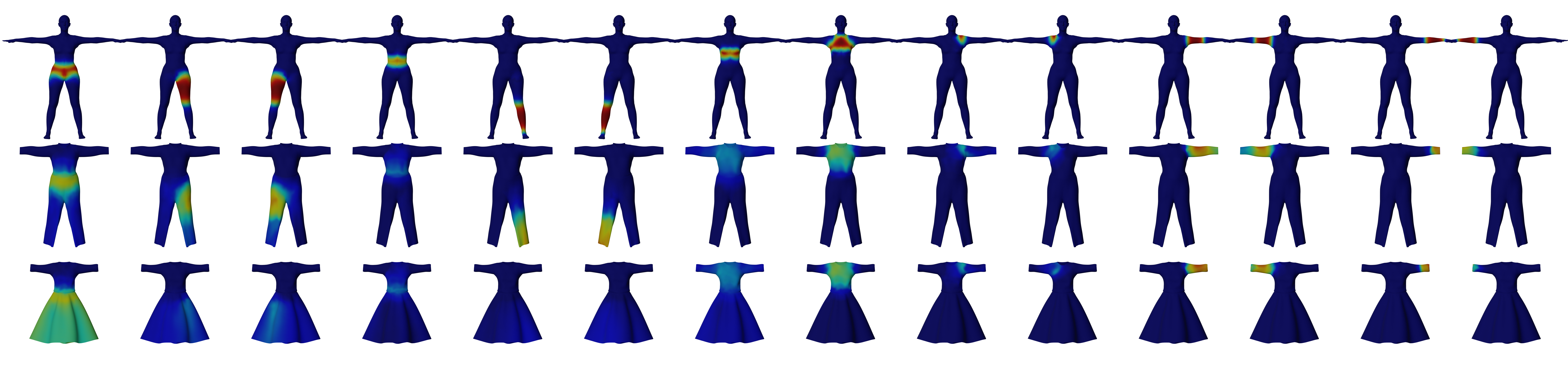}
    \caption{Blend weights comparison after training. First row: SMPL. Middle row: trouser-like. Last row: skirt-like.}
    \label{fig:blendweights}
\end{figure*}

We analyze the blend weights distribution for trouser-like garments (body homotopies) and skirt-like garments. Fig. \ref{fig:blendweights} shows the blend weights distribution for SMPL and two different garments (jumpsuit and dress). Note that we do not apply direct supervision to blend weights. Thus, the network predicts blend weights to minimize L2 loss only on final garment vertices. We can observe how the distribution of blend weights is very similar to those of SMPL. This supports the assumption that cloth closely follows body motion, since the network is able to learn by itself this distribution in order to minimize Euclidean error. Nonetheless, we observe a significant difference for skirts, where the predominant blend weights are the ones corresponding to the SMPL root joint. Skirts break the aforementioned assumption and the network learns to avoid relying on body motion by assigning skirts to root joint instead of leg joints.

\begin{table}[]
\centering
\begin{tabular}{lcccc}
\multicolumn{1}{c}{} & Euclidean error & \multicolumn{1}{l}{Edge} & \multicolumn{1}{l}{Bend} & \multicolumn{1}{l}{Collision} \\ \hline
Data & 36.73 & 1.16 & 0.015 & 15.4\% \\
Phys & 37.67 & 1.17 & 0.013 & 12.9\% \\ \hline
\end{tabular}
\caption{Results without PSD. First row corresponds to training with L2 loss on PBS data only. Second row is trained with L2 combined with $\mathcal{L}_{phys}$. We observe how performance is greatly compromised without PSD.}
\label{tab:no-psd}
\end{table}

In the main paper we assume it is known that garment animation on top of a human body cannot be done with skinning alone. First, since cloth behaviour is highly non-linear, skinning is not able to model it properly. Secondly, since SMPL itself has PSD, it would be impossible to achieve physically consistent predictions without PSD as well in the outfit. Nonetheless, to further prove this, we perform two additional experiments with no PSD (no $\mathbf{D}_{data}$ nor $\mathbf{D}_{phys}$). Tab. \ref{tab:no-psd} shows the obtained results. In the first experiment we train only with L2 loss against PBS data. On the second one, we apply both, L2 and $\mathcal{L}_{phys}$ losses. As it can be seen, the capacity of the model is highly compromised and, thus, Euclidean error is much higher than $\textit{DeePSD}$. Then, we observe that applying physical consistency loss has almost no effect, yielding a very high number of collisions. Nonetheless, since templates already have cloth consistency (edge and bend), we see how linear transformations (skinning) are able to maintain such constraints.

\section{Network Architecture and Training}

In this section with further detail network architecture and training process. Note that code is provided along with supplementary material to ease reproducibility and understanding of the model. In the main paper we describe the input as $\mathbf{T}\in\mathbb{R}^{N\times3}$. We empirically observed an slight increase in performance by concatenating vertex normals to each vertex. Therefore, the final input is defined as $\mathbf{T}\in\mathbb{R}^{N\times6}$. Then, $\Phi$ is defined as $4$ layers of graph convolutions with a perceptive field of $K = 1$ each and dimensionalities $32$, $64$, $128$ and $256$. Then, $\Phi$ also includes a fully-connected layer to compute global descriptor with dimensionality $256$. Global descriptor is concatenated with each vertex local descriptor to form per-vertex feature vectors with $F = 512$. Then, $\Omega$, $\Psi$ and $\chi$ are composed of $4$ fully-connected layers each, applied to vertices (vertices are \textit{samples}). Then, $\Omega$ has dimensionalities $128$, $64$, $32$ and $24$. $\Psi$ and $\chi$ have a similar architecture with dimensionalities $256$, $256$, $256$ and $P\times3$ (where $P = 128$ is the dimensionality of the high-level pose embedding $\Theta$). Finally, the MLP used for obtaining $\Theta$ from $\theta$ is composed of $4$ fully-connected layers with $256$ dimensions each, except the output layer, with dimensionality $P = 128$. We empirically observed that normalizing $\Theta$ as $\bar{\Theta_i} = \Theta_i / \sum_k\Theta_k$ heavily increases training stability.

We implemented our model using TensorFlow. We train each model for $10$ epochs. Starting with a batch size of $4$ and doubling it every $2$ epochs.  We use Adam optimizer with an initial learning rate of $0.001$. As stated in the main paper, it is not useful to train $\chi$ before the rest of the model has achieved convergence. Thus, we train $\chi$ independently during $10$ epochs after training the rest of the model ($\Phi$, $\Omega$ and $\Psi$). Regarding $\Omega$ without direct supervision, it will converge to the weights presented in Sec. \ref{sec:blendweights} of this document. Nonetheless, we empirically observed that it is not always guaranteed, most likely due to sensitivity to initialization, batch order or data bias. Thus, during the first epoch only, we apply a L2 loss based on a prior distribution. This prior relies again on the assumption that cloth follows the body. Then, $\Omega$ uses the nearest body vertex (in canonical pose) blend weights as labels. This ensures a correct initialization and slightly speeds up convergence. We further complement this prior with a prior on deformations. We assume that deformations are small. Again, for the first epoch only, we apply an L2 regularization on PSD.

\section{Performance}

We train our model in a GTX1080Ti for $10$ epochs, taking around $8$-$16$ hours (unsupervised losses are computationally expensive). Our model has a size of $4.4$Mb. For deployment, we run the model once per outfit to obtain animated 3D models in standard format. This format is highly computationally efficient due to exhaustive optimization of current graphics engines and GPUs. The only extra computational cost is obtaining the high-level pose embedding. Since an MLP is the most basic deep-learning model, it is unlikely to find a more efficient deep-based approach for 3D animation. In practice, we get $3$-$6$ms per sample and around $0.1$ms for batched samples, depending on vertex count, within TensorFlow pipeline. Proper integration into commercial graphics engine might further increase efficiency.

\end{document}